\documentclass[times, 10pt,twocolumn]{article}
\usepackage{latex8}
\usepackage{times}
\usepackage{balance}

\usepackage{graphics} 
\usepackage{epsfig} 

\newtheorem{theorem}{Theorem}[section]

\newtheorem{lemma}[theorem]{Lemma}
\newtheorem{proposition}[theorem]{Proposition}

\newenvironment{proof}[1][Proof]{\textbf{#1.} }{\ \rule{0.5em}{0.5em}}

\begin{document}

\title{Linear Time Recognition Algorithms for Topological Invariants in 3D}

\author{Li Chen\\
\emph{U of the District of Columbia}\\
\emph{lchen@udc.edu}\\
\and
 Yongwu Rong\\
\emph{George Washington University}\\
\emph{rong@gwu.edu}\\
}

\maketitle
\thispagestyle{empty}

\begin{abstract}
In this paper, we design  linear time algorithms to recognize and determine 
topological invariants such as the genus and homology groups in 3D. These properties 
can be used to identify 
patterns in 3D image recognition. 
This has many and is expected to have more applications in 3D medical image analysis. 
Our method is based on cubical images with direct adjacency, also called 
(6,26)-connectivity 
images in discrete geometry. According to the fact that there are
only six types of local surface points in 3D and a discrete version
of the well-known Gauss-Bonnett Theorem in differential geometry, we first 
determine the genus of a closed 2D-connected 
component (a closed digital surface). Then, we use Alexander duality to obtain 
the homology groups of a 3D object in 3D space.  This idea can be extended to 
general simplicial decomposed manifolds or cell complexes in 3D.
\end{abstract}


\Section{Introduction}

In recent years, there have been a great deal of new developments in applying topological tools to image analysis. In particular, computing topological invariants has been of great 
importance in understanding the shape of an arbitrary 2-dimensionall (2D) or 3-dimensional (3D) object~\cite {Kaz03}.
The most powerful invariant of these objects is the fundamental group~\cite{Hat}. 
Unfortunately,
fundamental groups are highly non-commutative and therefore difficult to work with. 
In fact, the general problem in determining whether two given groups are isomorphic 
is undecidable
(meaning that there is no algorithm can solve the problem)~\cite{Nov}. 
For fundamental groups
of 3D objects, this problem is decidable but no practical 
algorithm has been found yet.
As a result, homology groups have received the most attention because their 
computations are more feasible and they still provide significant information 
about the shape of the object~\cite{Day98} ~\cite{Kaz04} ~\cite{Dam08}. 
This leads to the motivating problem addressed in this paper: Given a 3D object in 
3-dimensional Euclidean space
$R^3$, determine  homology groups of the object in the most effective way by only 
analyzing  the 
digitization of the object.


The properties of   homology groups have applications in many areas of 
bioinformatics and image 
processing ~\cite{Kaz04}. We particularly look at a set of points in 3D digital space, and
our purpose is to find  homology groups of the data set.

Many researchers have made significant contributions in this area. For $R^3$, based on
simplicial  decomposition,  Dey and Guha have developed an algorithm 
for computing the homology group with generators
in $O(n^2 \cdot g)$, where $g$ is the maximum genus among all disconnected 
boundary surfaces. This algorithm has been improved by Damiand et al ~\cite{Dam08}. They used the boundary information to simplify the process. However,
the overall time complexity remained the same. 
 
In 2D, both $R^2$ and the cubical complex of linear algorithms (which is similar 
to that of digital spaces) are
found to calculate Betti numbers, which  are essentially the same as the genus  
\cite{Del95} and
\cite {Kaz03}. 

Other algorithms for homology groups in cubical spaces are studies in 2D, these algorithms are
either $O(n log^2 n )$  in \cite{Kot06} or   $O(n log^3 n )$ in \cite {Kaz03}.  
In general, the homology group can only be computed in $O(n^3)$ time for the cubical complex 
in \cite {Kaz03}. More about computational homology is discussed in  ~\cite{Kaz04}.

In this paper, we discuss the geometric and algebraic properties of manifolds 
in 3D digital spaces and the optimal algorithms for calculating these properties. 
We consider {\em digital manifolds}  as defined in ~\cite {Che04}. 
More information related to digital geometry and topology 
can be found in ~\cite{KR} and ~\cite{KoR}.

In this paper, we introduce an optimal algorithm with
time complexity $O(n)$ to compute genus and homology groups  in 3D digital space, 
where $n$ is the size of the input data.  
In Section 2, we review some properties of digital surfaces and manifolds ~\cite {Che04}.
Based on the classical Gauss-Bonnet Theorem, we calculate the genus of a digital 
closed surface in 3D. 
Section 3 covers the necessary background in 3-manifold topology. Using Alexander 
duality, we relate homology groups of a 3D object to its 2-dimensional boundaries. 
In Section 4,
we present our algorithm for homology groups.

\section{Gauss-Bonnet Theorem and Closed Digital Surfaces}
 
Cubical space with direct adjacency, or (6,26)-connectivity space, has the simplest 
topology in 3D digital spaces. It is also believed to be sufficient for the topological 
property extraction of 
digital objects in 3D. Two points are said to be adjacent in 
(6,26)-connectivity space if the 
Euclidean distance of these two points is 1, i.e., direct adjacency.
 
Let $M$ be a closed (orientable) digital surface in the 3D grid space in direct adjacency. 
We know that there are exactly 6-types of digital surface
points~\cite{Che04}\cite{CCZ99}.

\begin{figure}[h]
	\begin{center}

   \epsfxsize=2in 
   \epsfbox{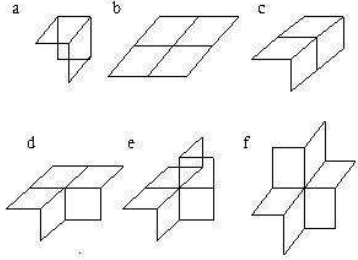}

	\end{center}
\caption{ Six types of digital surfaces points in 3D}
\end{figure}

Assume that $M_i$ ($M_3$, $M_4$, $M_5$, $M_6$) is the set of 
digital points with $i$ 
neighbors.   We have the following result for a simply connected 
$M$ ~\cite{Che04}:

\begin{equation}
          |M_3| =8 + |M_5| + 2 |M_6| .      
\end{equation}
 
\noindent $M_4$ and $M_6$ has two different types, respectively.

The Gauss-Bonnet theorem states that if $M$ is a closed manifold, then

\begin{equation}
          \int_{M} K_{G} d A = 2 \pi \chi(M)    
\end{equation}

\noindent where $d A$ is an element of area and $K_{G}$ is the Gaussian curvature. 

Its discrete form is

\begin{equation}
          \Sigma_{\{p \mbox{ is a point in } M\}} K(p) = 2 \pi \cdot (2- 2 g)  
\end{equation}


\noindent where $g$ is the genus of $M$.

Assume that $K_i$ is the curvature of elements in $M_i$, $i=$ 3,4,5,6. We have

\begin{lemma}\label{l21}
   (a) $K_3  = \pi/2$,
  (b) $K_4= 0$,  for both types of digital surface points,
  (c) $K_5  = - \pi /2$, and
   (d) $K_6  = - \pi$,  for both types of digital surface points.
\end{lemma}

\begin{proof}
       We can see that $K_4$ is always 0 since $K_G= K_1\times K_2$ and one of 
	the principal curvatures must be 0.  
We know that there exists a simply closed surface that contains only 8 points of 
$M_3$ and
several points within $M_4$. Based on (1) and (3), $8 \cdot K_3 = 2\pi\cdot (2-0)$. 
Therefore, $K_3= \pi/2$. There is a simply closed surface that only contains $M_3$, 
$M_4$, and $M_5$. 
So, $K_3=-K_5$. Same for $K_6=-2\cdot K_3$. So,  $K_6= - 2 \pi /2 = - \pi$.   
\end{proof} \\

Lemma 2.1 can also be calculated by the discrete Gaussian Curvature Theorem ~\cite{Pol02}.  
The curvature of the center point of the  polyhedra is determined by  
\begin{equation}
          \int_{M} K_{G} d A = 2 \pi - \Sigma_{i}{\theta_i}. 
\end{equation} 

\noindent Gauss-Bonnet Theorem will hold based on (4). 

Therefore, we can obtain the same results as Lemma 2.1. For example, in 3D digital space,
the angle of one face is $\pi/2$. So, $K_5=2 \pi - 5\cdot \pi/2 = - \pi/2$. 
(The proof of Lemma 2.1 is necessary if we are admitting the Gauss-Bonnet Theorem first
and then obtaining the curvature for each surface point. We include a proof here since 
the reference ~\cite{Pol02} does not contain such a proof.)

Given a closed 2D manifold, we can calculate the genus $g$ by counting the number of points in 
$M_3$, $M_5$, and $M_6$. According to (3), we have 

\[ \Sigma_{i=3}^{6} K_i \cdot |M_i| = 2 \pi \cdot (2- 2 g), \]
\[   \pi/2 \cdot |M_3| - \pi/2 \cdot |M_5| - \pi \cdot |M_6| = 2 \pi \cdot (2- 2 g), \]
\[     |M_3| -  |M_5| - 2 |M_6| = 4 \pi \cdot (2- 2 g). \]

\noindent Therefore,

\begin{equation}
           g = 1+ (|M_5|+2 \cdot |M_6| -|M_3|)/8. 
\end{equation} 

\begin{lemma}\label{l22}
 There is an algorithm that can calculate the genus of $M$ in linear time.

\end{lemma}

\begin{proof}
   Scan through all points (vertices) in $M$ and count the neighbors of each point. We can see
that a point in $M$ has 4 neighbors indicating that it is in $M_{4}$ as are $M_5$ and $M_6$.
 Put points to each category of 
 $M_i$. Then use formula (5) to calculate the genus $g$.
\end{proof} \\  

The two following examples show that the formula (5) is correct. The first example shown in 
Fig. 2, is the easiest case. In Fig.2 (a), there are 8 points in $M_3$ and no points in $M_5$ 
or $M_6$. (To avoid the conflict between a closest digital surface and a 3-cell ~\cite{Che04}, 
we can insert some $M_4$ points on the surface but not at the center point.)  
According to (5), $g=0$. Extend Fig. 2 (a) to a genus 1 surface as shown in Fig. 2 (b) where
there are still 8 $M_3$ points but 8 $M_5$ points. Thus, Fig. 2 (b) satisfies equation (5).
We can extend it to Fig. 2 (c), it has 16 $M_5$ points and 8 $M_3$ points. So $g=2$. 
Using the same method, one can insert more handles. 

The second example came from the Alexander horned sphere. See Fig. 3.  First we show a ``U''
shape base in Fig. 3(a). It is easy to see that there are 12 $M_3$ points and 4 $M_5$ points. So 
$g=0$ according to Equation (5). Then we attach a handle to Fig. 3(a) shown in Fig. 3(b). 
We have added 4 $M_3$ points and 12 $M_5$ points. $g = 1+ (|M_5|+2 \cdot |M_6| -|M_3|)/8=
1+(4+12-12-4)=1$. Finally, we add another handle to the other side of the ``U'' shape in Fig. 3(a), the genus number
increases by one since we still add  4 $M_3$ points and 12 $M_5$ points shown in Fig. 3(c).
$g=2$ for (c).
For more complex cases like the Alexander horned sphere, one just needs to insert two smaller handles to an existing handle, so the genus will increase accordingly.

\begin{figure}[h]
	\begin{center}

   \epsfxsize=2in 
   \epsfbox{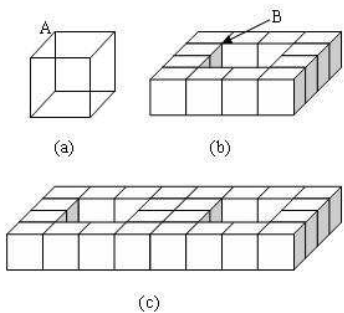}

	\end{center}
\caption{ Simple examples of closed surfaces with $g=0,1,2$}
\end{figure}

\begin{figure}[h]
	\begin{center}

   \epsfxsize=2.3in 
   \epsfbox{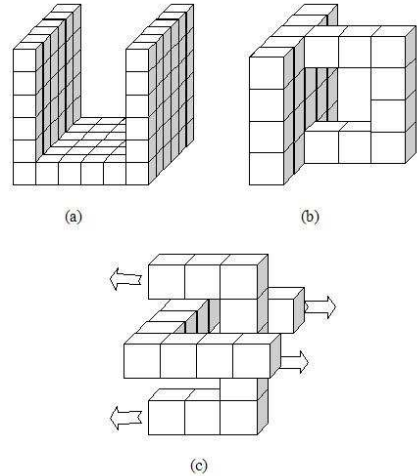}

	\end{center}
\caption{An example came from Alexander horned sphere in digital space}
\end{figure}

The above idea can be extended to simplicial cells (triangulation) or even general CW $k$-cells.
This is because, for a closed discrete surface, we can calculate Gaussian curvature at each 
vertex point using formula (4). (The key is to calculate all angles separated by 1-cells at
a vertex) Then use (3) to obtain the genus $g$. 
Since each line-cell (1-cell) is involved in exactly two 2-cells, it is only associated with four angles. Therefore the total complexity will be  
 $O(|E|)$ where $E$ is the set of 1-cells (edges).  Thus,

\begin{lemma}\label{l22}
 There is an algorithm that can calculate the genus of a closed simplicial surface in
  $O(|E|)$ where $E$ the set of 1-cells (edges).
\end{lemma}

There are examples that $|E|$ is not linear to the number of vertices $|V|$.

\section{Homology Groups of Manifolds in 3D Digital Space} 

Homology groups are other invariants in topological classification.  For a $k$-manifold ,
Homology group $H_{i}$, $i=0,...,k$ indicates the number of holes in each $i$-skeleton of the
manifold. Once we obtain the genus of a closed surface, we can then
calculate the homology groups corresponding to its 3-dimensional
manifold. 

Consider a compact 3-dimensional manifold in $R^3$ whose boundary is represented by a surface.
 We show its homology groups can be expressed in terms of its boundary surface (Theorem \ref{Jordan2}).
  This result follows
from standard results in algebraic topology. Since it does not seem to be explicitly stated or proved in any standard
reference, we include a self-contained proof here  \cite{Hat}.
This result follows from standard results in algebraic topology. It
also appears
in \cite{Day98} in a somewhat different form. For the convenience of readers,
we include a
short self-contained proof here.

First, we recall some standard concepts and results in topology.
Given a topological  space $M$, its homology groups, $H_i(M)$, are certain measures of $i$-dimensional "holes" in $M$.
For example, if $M$ is the solid torus, its first homology group $H_1(M)\cong Z$, generated by its 
longitude which goes around the obvious hole. For a precise definition, see, e.g. \cite{Hat}.
Let $b_i = \mbox{rank} H_i(M, Z)$ be the $i$th Betti number of $M$. The Euler characteristic of $M$ is defined by

\[ \chi (M) = \sum_{i\geq 0} (-1)^i b_i \]

If $M$ is a 3-dimensional manifold, $H_i(M)=0$ for all $i>3$ essentially because 
there are no $i$-dimensional holes.
Therefore, $\chi (M)  = b_o - b_1 + b_2 - b_3$.
 Furthermore, if $M$ is in
$R^3$, it must have nonempty boundary. This implies that $b_3 = 0$.

The following lemma is well known for 3-manifolds. It holds, with the same proof,
 for any odd dimensional manifolds.

\begin{lemma}\label{chi} Let $M$ be a compact orientable 3-manifold (which may or may not be in $R^3$).
\item{(a)} If $M$ is closed (i.e. $\partial M = \emptyset$), then $\chi (M) = 0$.
\item{(b)} In general, $\chi(M) = \frac{1}{2} \chi(\partial M)$.
\end{lemma}

\noindent
\begin{proof}
(a) If $\partial M = \emptyset,$ the result follows from
 the Poincare duality which says that $H_i(M)\cong H^{3-i}(M)$, and
 the Universal Coefficient Theorem which says that the free part of $H^{3-i}(M)$ is isomorphic to the free part
of $H_{3-i}(M)$. Together, they imply that $b_i=b_{3-i}$.
Hence $\chi(M) = b_0 - b_1 + b_2 - b_3=0$.

(b) In general, we consider the {\it double} of $M$ denoted by $DM$, which is obtained by gluing two copies of $M$ along
$\partial M$ via the identity map. By what we just proved, $\chi(DM)=0$. On the other hand, 
the Euler characteristic satisfies a nice additive property:
$\chi(DM) = \chi(M) + \chi(M)-\chi(\partial M)= 2\chi(M)-\chi(\partial M)$. This implies
$\chi(M) = \frac{1}{2} \chi(\partial M)$.
\end{proof} \\

Next, we recall the Alexander duality.

\begin{proposition}
Let $X \subset S^n$ be a  compact, locally contractible subspace of $S^n$ where $S^n$ is
 the $n$-dimensional sphere.
 Then \\
$\tilde{H}_i(S^n -X) \cong \tilde{H}^{n-i-1}(X)$ for all $i$ where $\tilde{H}$ is the reduced homology.
\end{proposition}

We remark that $S^n$ is the one point compactification of $R^n$. Therefore, a submanifold in $R^n$ is automatically
considered as a submanifold in $S^n$ in a natural way.
Conversely, a submanifold $M$ in $S^n$ is automatically a submanifold in $S^n$ unless $M=S^n$.

Before we prove Theorem \ref{Jordan2}, we first prove a special case when $\partial M$ is connected.

\begin{lemma}\label{Jordan}
Let $S$ be a closed connected surface in $S^3$. \\
(a) Its complement, $S^3 - S$, must have exactly two connected components. We denote them by $M$ and $M'$. \\
 (b) $H_1(M)\cong H_1(M')\cong Z^{\frac{1}{2} b_1(S)}$,
$H_2(M)\cong H_2(M')=0$.
\end{lemma}

\noindent
\begin{proof}
(a) By the Alexander duality, $\tilde{H}_0(S^3 - S) \cong H^2(S) \cong Z$. Therefore $S^3 - S$ must have exactly two components.

(b) Again, by the Alexander duality, $H_2(S^3-S)\cong \tilde{H}^0(S) =0$, and this implies $H_2(M)\cong H_2(M')=0$.
We also have $H_1(M)\oplus H_1(M')\cong H_1(M\sqcup M')= H_1(S^3-S)\cong H^1(S)\cong Z^{b_1(S)}$.
In particular, there is no torsion in $H_1(M)$ or $H_1(M')$.
By Lemma \ref{chi},
$1-\mbox{rank} H_1(M) = 1-\mbox{rank} H_1(M') = \frac{1}{2} \chi(S) = \frac{1}{2} (2-b_1(S))$.
This implies $H_1(M)\cong H_1(M)\cong Z^{\frac{1}{2} b_1(S)}$
\end{proof} \\

Now we consider a general compact connected 3-manifold $M$ in $S^3$.
Its boundary, $\partial M$, is a closed orientable 2-dimensional manifold possibly with
several components.

\begin{theorem}\label{Jordan2}
Let $M$ be a compact connected 3-manifold in $S^3$. Then
\begin{enumerate}
\item [(a)]  $H_0(M)\cong Z$.
 \item [(b)] $H_1(M) \cong Z^{\frac{1}{2} b_1(\partial M)}$, i.e. $H_1(M)$ is torsion-free with rank being half of rank $H_1(\partial M)$.
\item[(c)] $H_2(M) \cong Z^{n-1}$ where  $n$ is the number of components of $\partial M$.
\item[(d)] $H_3(M)=0$ unless $M=S^3$.
\end{enumerate}
\end{theorem}

\noindent
\begin{proof}
Statements (a) and (d) are obvious: (a) follows from connectedness of $M$, (d) is due to the fact that
$\partial M$ is non-empty. Next, we prove (c) and (d).

Let $S_1, \cdots, S_n$ be the connected components of $\partial M$.
By Lemma \ref{Jordan}, each $S_i$ separates $S^3$ into two connected components, $M_i$ and $M'_i$. Since $M$ is connected, it must be entirely contained in either $M_i$ or $M'_i$.
Let $M'_i$ be the one containing $M$. It follows that
 $S^3 - \cup_i S_i = M \sqcup \sqcup_i M_i$.

By the Alexander duality, $H_2(M\sqcup \sqcup_i M_i) = H_2(S^3 - \cup_i S_i)
\cong \tilde{H}^0(\cup_i S_i) \cong Z^{n-1}$.
 But $H_2(M_i)=0$ for each $i$ (Lemma \ref{Jordan}). Therefore
 $H_2(M) \cong H_2(M\sqcup \sqcup_i M_i) \cong
  Z^{n-1}$.

Next, also by the Alexander duality, $H_1(M\sqcup \sqcup_i M_i) = H_1(S^3 - \cup_i S_i)
\cong H^1(\cup_i S_i) = H^1(\partial M) \cong Z^{b_1(\partial M)}$.
But LHS = $H_1(M) \oplus \oplus_i H_1(M_i) = H_1(M) \oplus \oplus_i Z^{\frac{1}{2} b_1(S_i)}
\cong H_1(M) \oplus Z^{\frac{1}{2} b_1(\partial M)}$.
It follows that
$H_1(M)\cong Z^{\frac{1}{2} b_1(\partial M)}$.
\end{proof} \\

\section{A Linear Algorithm of finding Homology Groups in 3D}

Based on the results we presented in Sections 2 and 3, we
now describe  a linear algorithm for computing the homology group of 3D objects
in 3D digital space.

Assuming we only have a set of points in 3D. We can digitize this set into 3D digital spaces. 
There are two ways of doing so: (1) by treating each point as a cube-unit that is called the 
raster space, 
(2) by treating each point as a grid point. It is also called the point space. 
These two are dual spaces.
Using the algorithm described in ~\cite{Che04}, we can determine whether the digitized set forms a 
3D manifold in 3D space in direct adjacency for connectivity. The algorithm is in linear time. 

The more detailed considerations of recognition algorithms related to 3D manifolds can be 
found in ~\cite{BK08} where
Brimkov and Klette made extensive investigations in boundary tracking.  The discussions 
of 3D objects in raster space can be found in ~\cite{Lat}. \\

\noindent {\bf Algorithm 4.1} Let us assume that we have a connected $M$ that 
is a 3D digital manifold in 3D.

\begin{description}

 \item [Step 1.] Track the boundary of $M$, $\partial M$, which is a union of several closed surfaces. 
This algorithm only needs to scan though all the points in $M$ to see if 
the point is linked to  
a point outside of $M$. That point will be on boundary. 

\item [Step 2.]  Calculate the genus of each closed surface in $\partial M$ using the method 
described in Section 2. We just need to count the  number of neighbors on a surface.  
and put them in $M_i$, using the formula (5) to obtain $g$.

\item [Step 3.] Using the Theorem \ref{Jordan2}, we can get $H_0$, $H_1$, $H_2$, and $H_3$. 
                $H_0$ is $Z$. For $H_1$, we need to get $b_1(\partial M)$ that is just 
				the summation of the genus in all connected components in  $\partial M$. (See \cite{Hat}
				and \cite{Day98}.) 
				$H_2$ is the number of components in $\partial M$. $H_3$ is trivial.
\end{description}

\begin{lemma}
           Algorithm 4.1 is a linear time algorithm. 
\end{lemma}

\begin{proof}
       Step 1 uses linear time. We can first track all points in the object using breadth-first-search.
We assume that the points in the object are marked as ``1'' and the others are marked as ``0.''
Then, we test if a point in the object is adjacent to both ``0'' and ``1'' by using 26-adjacency for linking to ``0.''  
Such a point is called a boundary point. It takes linear time because the total number of adjacent points is only 26. 
Another algorithm is to test if each line cell on the boundary has exactly two parallel moves on the boundary ~\cite{Che04}. This procedure only takes linear time for the total number of boundary points in most cases. 

Step 2 is also in linear time by Lemma 2.2. 

Step 3 is just a simple math calculation. 
For $H_0$, $H_2$, and $H_3$, they can be computed in constant time. For $H_1$, 
the counting process is at most linear. 
\end{proof} \\

Therefore, we can use linear time algorithms to calculate $g$ and all homology
 groups for digital manifolds in 3D based on Lemma 2.2 and Lemma 4.1.

\begin{theorem}
           There is a linear time algorithm to calculate all homology
 groups for each type of manifolds in 3D.  
\end{theorem} 

To some extent, researchers are also interested in space complexity that is regarded 
to running space needed beyond the input data. Our algorithms do not need to store 
the past information, the algorithms presented
in this paper are always $O(\log n)$. Here, $\log n$ is the bits 
needed to represent a number $n$.

{\em Acknowledgement.} The authors would like to thank Professor Allen Hatcher for 
getting the authors connected which led to the result of this paper. The second author 
is partially supported by NSF grant DMS\-051391.








\end{document}